# Towards Fair Knowledge Transfer for Imbalanced Domain Adaptation

Taotao Jing , Bingrong Xu , and Zhengming Ding, *Member, IEEE*

*Abstract*—Domain adaptation (DA) becomes an up-and-coming technique to address the insufficient or no annotation issue by exploiting external source knowledge. Existing DA algorithms mainly focus on practical knowledge transfer through domain alignment. Unfortunately, they ignore the fairness issue when the auxiliary source is extremely imbalanced across different categories, which results in severe under-presented knowledge adaptation of minority source set. To this end, we propose a Towards Fair Knowledge Transfer (TFKT) framework to handle the fairness challenge in imbalanced cross-domain learning. Specifically, a novel cross-domain knowledge propagation technique is proposed with the guidance of within-source and cross-domain structure graphs to smooth the manifold of the minority source set. Besides, a cross-domain fulfillment augmentation strategy is exploited achieve domain adaptation. Moreover, hybrid distinct classifiers and cross-domain prototype alignment are adopted to seek a more robust classifier boundary and mitigate the domain shift. Such three strategies are formulated into a unified framework to address the fairness issue and domain shift challenge. Extensive experiments over two popular benchmarks have verified the effectiveness of our proposed model by comparing to existing state-of-the-art DA models, and especially our model significantly improves over 20% on two benchmarks in terms of the overall accuracy.

## I. INTRODUCTION

**M**ASSIVE amounts of well-annotated training data contribute to a great deal to the development of deep neural networks (DNNs) and the significant performance improvement on image recognition tasks [1]–[4]. Unfortunately, the soaring cost of collection and annotation makes it difficult to gather sufficient data to train effective DNNs in some real applications. Domain Adaptation (DA) casts a light on the problem by borrowing knowledge from another well-annotated but different distributed domain [5]–[10].

Domain adaptation aims to mitigate the distribution difference across different domains and obtain a model designed on the well-labeled source domain but applicable to the unlabeled target domain. Existing domain adaptation efforts consist of two categories, minimizing distribution discrepancy and adversarial training mechanism. The first one, minimizing the distribution discrepancy, seeks domain invariant representation, where Maximum Mean Discrepancy (MMD) is one of the crucial members [11], [12]. The other group of methods, adversarial training mechanism, performs an adversarial optimization strategy to train a domain discriminator while optimizing a feature generator to map both source and target domain data into a shared latent space [13]. Besides aligning the source and target domain-level distribution shifts, some works reveal the task-specific decision boundaries to explore the potential of class-wise adaptation [5], [13]. However, current domain adaptation efforts all assume an ideal situation that the annotated source domain data are sufficient and balanced distributed across all categories, which is hard to guarantee. The long-tail distribution is a more realistic and practical situation in data collection [14]–[16].

To solve the distribution imbalance, some researchers focus on the Imbalanced Domain Adaptation task, where they seek to mitigate the distraction of the imbalance by assigning an importance weight to each sample [8], [17]–[21]. However, such strategies still rely on the importance weights assigned to the target domain samples by the source classifier, which could be undependable when in some extreme situation when some categories even lack enough data to train a reasonable classifier. Under such extreme situation, the ability to maintain the performance on specific categories with limited training data is crucial and challenging, leading to transfer fairness problems [22]. Conventional domain adaptation methods or works focusing on imbalanced domain adaptation tasks ignore those classes lacking sufficient training data resulting in domination on those many-shot categories and unreliable on the few-shot classes.

In this paper, we consider the source fairness challenge in domain adaptation under the extreme condition when the available source domain is extremely imbalanced as illustrated in Fig. 1, i.e., some source domain categories only contain a few labeled samples for training. Consequently, we propose a novel Towards Fair Knowledge Transfer (TFKT) framework to guarantee faithful cross-domain adaptation. The contributions of this work are summarized in four folds as follows:

- First of all, we propose the knowledge propagation within the source domain and across source and target domains with the weighted structure graph guidance to smooth the manifold and alleviate the distraction caused by the undesired random few-shot samples belonging to the source domain minority categories.
- Secondly, we exploit the cross-domain fulfillment augmentation strategy based on the refined data through

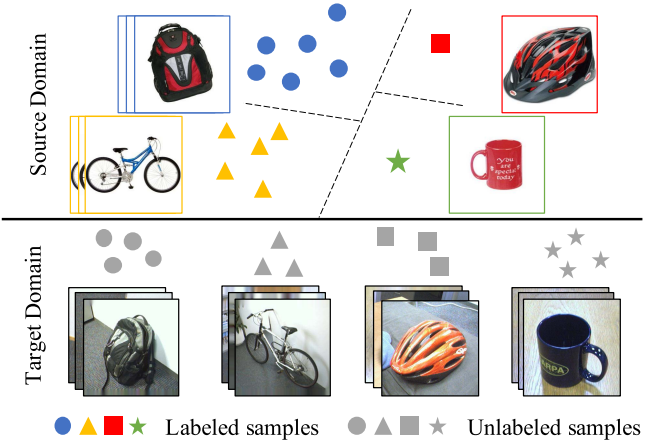

Fig. 1. Illustration of the imbalanced domain adaptation task. The colored shapes denote labeled but extremely imbalanced source domain data, in which some categories only contain few samples, e.g., one-shot. The gray shapes are unlabeled target domain data.

knowledge propagation to achieve across-domain alignment and eliminate the domain shift.

- Thirdly, we enhance the faithful knowledge transfer by exploring dual-classifier mechanism and cross-domain alignment, to seek more robust task-specific classification boundaries and domain-invariant feature representation.
- Finally, we extensively evaluate our proposed model under various challenging but realistic settings and compare the performance with state-of-the-art methods. The superior results, even in the extreme situation with only one labeled source sample available for some classes, emphasize the effectiveness of our model.

## II. RELATED WORK

In this section, we briefly review the most recent studies of domain adaptation as well as learning strategies handling data distribution imbalance issues.

### A. Domain Adaptation

Unsupervised domain adaptation (UDA) aims to resolve the lack of well-labeled data for training via bridging the knowledge gap between an external source domain with sufficient annotation and the unlabeled target domain [23]–[30]. A representative solution to domain adaptation challenges is reducing the domain distribution discrepancy between the source and target domains [24]. Minimizing the maximum mean discrepancy (MMD) [25] is one typical and sufficiently explored application of this strategy, which can also be implemented as multi-kernel MMD with a domain adaptation network (DAN) [31]. However, these efforts only align the source and target domain-wise distribution shift, while Weighted Domain Adaptation Network (WDAN) [27] reweighs the source domain each category weights through introducing class-specific auxiliary weights into the original MMD metric. Moreover, many works evaluate the marginal and conditional distribution importance for cross-domain alignment [30], [32], [33]. Recently, adversarial learning attracted attention on domain adaptation space [34], [35], by seeking a domain-invariant feature generator and binary domain confusion discriminator.

However, most methods proposed to address UDA problem only consider that the source and target domains share an identical label space, which is always satisfied in real-life and decreases the feasibility in practice. Thus, various domain adaptation variants are explored recently breaking the constraint of identical label space across domains. Along this line, Partial domain adaptation (PDA) focuses on the situation when the source domain label space subsumes the target domain [18], [36], [37]. On the contrary, open-set domain adaptation (OSDA) assumes some unknown categories exiting in the target domain [10], [38], [39]. Moreover, universal domain adaptation (UniDA) tackles the generalized setting without any prior knowledge on the label sets [40], [41]. Efforts on such generalized settings shed light on designing practical domain adaptation models.

### B. Fairness Transfer

Different from assuming adequate samples available for training as standard DA assumption, fairness arises in domain adaptation tasks when the training source domain data is distributed extremely imbalanced. Lack of sufficient training data from specific categories will mislead the model overfitting to those classes with adequate training samples. Some existing efforts manage the distribution imbalance problems by re-weighing the domain alignment loss [27], [42]. [43] addresses the fairness problems through involving adversarial mechanisms to optimize the fairness metrics, while [44] focuses on constraining the optimization of fairness objectives. Moreover, [45] aims to solve the fairness domain adaptation when attribute labels are not available in the domains, and [46] reveals that transfer learning can boost the accuracy of target tasks at the cost of lowering prediction fairness. However, such strategies tend to fail in extreme situations when only few samples are accessible in some categories, since their learned models will reflect those undesired biases from the training data distribution. Differently, we use cross-domain generation strategy to augment imbalanced source data, and hybrid classifier framework to enhance the model ability.

## III. THE PROPOSED METHOD

### A. Preliminaries and Motivation

In this paper, we consider a challenging but realistic domain adaptation task involving a well-labeled but extremely-imbalanced source domain dataset $\{\mathbf{X}_s, \mathbf{Y}_s\} = \{(\mathbf{x}_s^i, \mathbf{y}_s^i)\}_{i=1}^{n_s}$ and an unlabeled target domain dataset $\{\mathbf{X}_t\} = \{\mathbf{x}_t^j\}_{j=1}^{n_t}$, where $n_s$ and $n_t$ are the numbers of samples in the source and target domains, respectively. The source domain, denoted as $\mathcal{D}_s$, consists of two subsets, a majority-set $\mathcal{D}_s^m$ and a minority-set $\mathcal{D}_s^f$, where the majority-set $\mathcal{D}_s^m$ have $n_s^m$ samples available and each category consists of sufficient instances with annotations from the label space $\mathcal{Q}^m$, while the minority-set $\mathcal{D}_s^f$ only contains $n_s^f$ data drawn from $P$ categories from label space $\mathcal{Q}^f$, with limited $Q$ samples per class, and we describe it as $P$ way $Q$ shot task. $\mathbf{Z}_{s/t} = E(\mathbf{X}_{s/t}), \mathbf{Z}_{s/t} \in \mathbb{R}^{n_{s/t} \times d}$ is the source/target embedding representations extracted from pretrained backbone network $E(\cdot)$, where $d$ is the embedding dimension. In the rest of this work, we choose ResNet-50 [47] pretrained on ImageNet [48] as the convolutional backbone and accept the output before the last fully-connected layer as the embedding representation. The source and target domains are drawn from different distributions $\mathbf{P}_s$ and $\mathbf{P}_t$, but lie in the same label space $\mathcal{Q} = \mathcal{Q}^m \cup \mathcal{Q}^f$, thus the number of categories in the source and target domains are identical.

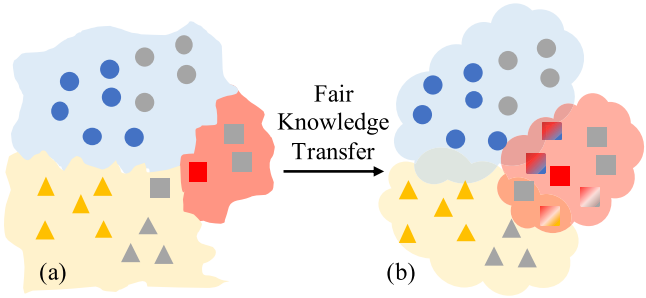

Fig. 2. Illustration of the fair knowledge transfer process, which is able to expand the feature space of categories with a few labeled samples, and balance the decision boundaries. Different shapes represent different categories, while colored and gray instances denote labeled source and unlabeled target samples, respectively.

Some source domain categories only contain a few samples for training that will fail the conventional unsupervised domain adaptation solutions, relying on plenty of well-labeled source domain data for supervised training. The extremely imbalanced distribution will distract the optimization direction and mislead the model overfitting to those classes with adequate training data while ignoring the few-shot categories. We address such challenges by refining the embedding representations of samples from few-shot categories through structure guided knowledge propagation to eliminate the undesired noise distraction, then synthesizing those few-shot categories to expand the feature distribution space and avoid the imbalance issues, and finally exploring the dual classifier mechanism to align the source and target domain and alleviate domain shift. The fair knowledge transfer process can expand the feature space of minority set categories with few labeled training samples, and smooth the decision boundaries (Fig. 2).

### B. Towards Fair Knowledge Transfer

We first present an overview of our proposed framework (Fig. 3). Firstly, the source domain few-shot category sample ($\mathbf{z}_s^f$) is augmented to obtain refined embeddings ($\tilde{\mathbf{z}}_{s/o}^f$) through the knowledge propagation within the source and across domains under the guidance of weighted structure graphs ($\mathbf{H}_{s/o}$). Besides, the refined synthesized samples ($\tilde{\mathbf{z}}_{s/o}^f$) are used to generate more synthesized instances ($\hat{\mathbf{z}}_m^f$) to expand the feature space and fulfill the distribution gap across domains through random combination. Finally, all the real and synthesized instances are mapped into a domain invariant space through the feature generator $F(\cdot)$, denoting the output features as $\mathbf{f} = F(\mathbf{z})$, guided by the discriminative cross-domain alignment and source domain supervision objectives obtained from the dual-classifier mechanism, consisting of a multi-layer neural network classifier $C_N(\cdot)$ and a prototypical classifier

$C_P(\cdot)$. We will discuss details about each part in the following sections.

### C. Cross-Domain Feature Augmentation (CDA)

*1) Data Augmentation Through Embedding Propagation:* The most challenging difference between the extreme imbalanced domain adaptation and conventional domain adaptation tasks is that some categories from the source domain only contain few labeled samples for training. The optimization process of previous domain adaptation solutions would be dominated and misled by the majority set classes having sufficient training data and fail on the minority set categories. Besides, the limited labeled samples from the minority set categories may lie far from the class center in the feature space, which cannot represent the corresponding categories distribution characteristic. To address this challenge, we first explore the augmentation strategy *embedding propagation* within the source domain [49]. Specifically, for each source sample from the minority set $\mathcal{D}_s^f$, an interpolated embedding is constructed through the combination of its neighbors with the knowledge propagated under the guidance of a weighted graph. The goal of embedding propagation is to remove the noise from the features and smooth the embedding manifold, which will benefit the generalization and effectiveness of semi-supervised learning methods [49]–[51].

Firstly, we build a similarity adjacency matrix $A_s \in \mathbb{R}^{n_s \times n_s}$ for all samples in the source domain $\mathcal{D}_s$, and each element in $A_s$ is computed as:

$$A_s^{i,j} = \exp(-\frac{d_{ij(s)}^2}{\sigma_s^2}) = \exp(-\frac{\|\mathbf{z}_s^i - \mathbf{z}_s^j\|^2}{\sigma_s^2}), \qquad (1)$$

in which $d_{ij(s)} = \|\mathbf{z}_s^i - \mathbf{z}_s^j\|_2$ is the distance between two samples features $\mathbf{z}_s^i$ and $\mathbf{z}_s^j$, both from the source domain $\mathcal{D}_s$, and $\sigma^2$ is the scaling factor which is set as the standard variation of the distances, i.e., $\sigma^2 = Var(d_{ij(s)}^2)$, and $A_s^{kk} = 0, \forall k$ [52]. Then based on the pair-wise similarity adjacency graph, the Laplacian matrix can be obtained as:

$$L_s = D_s^{-1/2} A_s D_s^{-1/2}, \qquad (2)$$

where $D_s^{ii} = \sum_j A_s^{ij}$. Then based the propagator proposed in [53], the weighted knowledge propagation graph within the source domain, denoted as $H_s$, can be calculated as:

$$H_s = (\mathrm{I} - \alpha L_s)^{-1}, \qquad (3)$$

where $\alpha \in \mathbb{R}$ is a scaling factor which is fixed as 0.2 following [49], and I is the identity matrix.

Then for each sample $\mathbf{z}_s^i$ from the source minority set $\mathcal{D}_s^f$, an interpolated embedding $\tilde{\mathbf{z}}_s^i$ is constructed by the structure knowledge propagated from all its neighbors under the guidance of the weighted propagation graph:

$$\tilde{\mathbf{z}}_s^i = \sum_{\mathbf{x}_s^j \in \mathcal{D}_s^f} H_s^{ij} \mathbf{z}_s^j, \qquad (4)$$

in which $\tilde{\mathbf{z}}_s^i$ share the same label as $\mathbf{z}_s^i$, and the augmented interpolated embeddings for all source domain minority set samples make up the set $\tilde{\mathcal{D}}_s^f = \{\tilde{\mathbf{z}}_s^i | \mathbf{x}_s^i \in \mathcal{D}_s^f\}$ lying in the identical label space as $\mathcal{D}_s^f$. Since the constructed $\tilde{\mathbf{z}}_s^i$

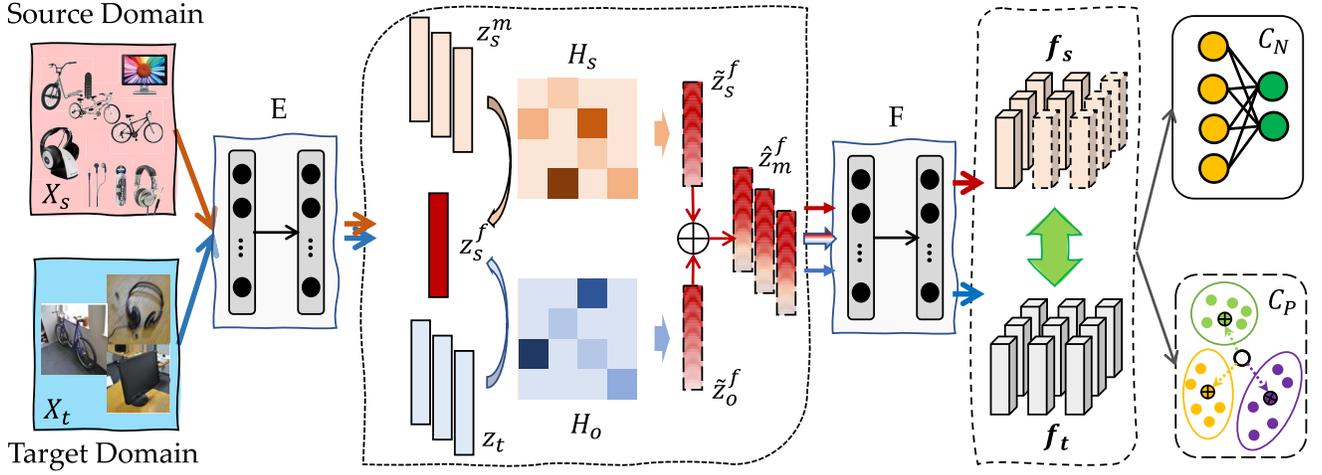

Fig. 3. Overview of our proposed framework, where both source and target raw images $(\mathbf{x}_{s/t})$ are input to pretrained deep convolutional neural networks $E(\cdot)$ to extract embedding representations $(\mathbf{z}_{s/t})$. $\mathbf{z}_s^m$ denotes the samples from the majority set categories, while $\mathbf{z}_s^f$ denotes the samples from the source domain minority categories, and $\mathbf{z}_t$ are the samples from the target domain without labels. $\mathbf{H}_s$ is the weighted graph illustrating the structure of the source domain data, while graph $\mathbf{H}_o$ is obtained based on the relationships between samples from the source domain few-shot set and the target domain samples. $\tilde{\mathbf{z}}_{s/o}^f$ are the augmented refined embeddings through knowledge propagation within the source domain and across domains, respectively, which are used to generate more synthesized instances $(\hat{\mathbf{z}}_m^f)$ through random combination to expand the feature space and fulfill the distribution gap across domains. All real and synthesized embedding instances are mapped into a domain invariant space through the feature generator $F(\cdot)$, denoting the output features as $\mathbf{f}_{s/t}$. The dual-classifier scheme consists of multi-layer neural network classifier $C_N(\cdot)$ and prototypical classifier $C_P(\cdot)$, which aims to preserve source supervision.

involves the structure information from the whole source domain, so such knowledge propagation augmented samples can expand the corresponding category feature space and eliminate the undesired noise from outliers.

*2) Cross-Domain Knowledge Propagation:* Except labeled source domain data, the unlabeled target domain is also rich in the structure information corresponding to the source domain. However, due to the domain shift, which is one of the main challenges in domain adaptation tasks caused by the different data distribution across domains, we cannot directly put the source and target domain data together and apply the knowledge propagation globally. Because the knowledge propagation graph computed overall source and target domain data will be dominated by the relationship between samples from the same domain, while the structure knowledge between samples across domains is easy to be ignored compared to the close relationship within source/target domain. Based on this, instead of directly combining the source and target domain together, we propose the cross-domain knowledge propagation from the target domain $\mathcal{D}_t$ to the few-shot source domain minority set $\mathcal{D}_s^f$.

Specifically, by putting the source domain minority set $\mathcal{D}_s^f$ and the target domain data $\mathcal{D}_t$ together making up a new dataset $\mathcal{D}_o = \mathcal{D}_s^f \cup \mathcal{D}_t$. The adjacency matrix $A_o \in \mathbb{R}^{(n_s^f + n_t) \times (n_s^f + n_t)}$ based on the dataset $\mathcal{D}_o$ is computed as:

$$A_o^{i,j} = \exp(-\frac{d_{ij(o)}^2}{\sigma_o^2}) = \exp(-\frac{\|\mathbf{z}_o^i - \mathbf{z}_o^j\|^2}{\sigma_o^2}), \quad (5)$$

where $d_{ij(o)}^2 = \|\mathbf{z}_o^i - \mathbf{z}_o^j\|^2$ is the distance between the samples $\mathbf{z}_o^i$ and $\mathbf{z}_o^j$, both from $\mathcal{D}_o$, the scaling factor $\sigma_o^2 = Var(d_{ij(o)}^2)$, and $A_o^{kk} = 0, \forall k$ [52]. It is noteworthy that due to the different distribution between the source and target domain, only considering the relationship between the given source domain minority-set sample and the target domain may mislead the structure knowledge in the adjacency matrix. So $A_o$ keeps the structure knowledge about the corresponding relationships among the source domain samples from minority set in $\mathcal{D}_s^f \subset \mathcal{D}_o$. In other words, $A_o$ contains both within-source and source-target structure information.

Similarly, the Laplacian matrix of $A_o$ is computed as:

$$L_o = D_o^{-1/2} A_o D_o^{-1/2}, \quad (6)$$

where $D_o^{ii} = \sum_j A_o^{ij}$, and the weighted cross-domain knowledge propagation graph is calculated as:

$$H_o = (I - \alpha L_o)^{-1}, \quad (7)$$

in which $\alpha$ is fixed as 0.2 following [49], same as Eq. 3.

Based on the cross-domain knowledge propagator $H_o$, for each sample $\mathbf{z}_s^i$ from the source domain minority set $\mathcal{D}_s^f$, a synthesized embedding $\tilde{\mathbf{x}}_o^i$ is constructed by the combination of its neighbors in the set $\mathcal{D}_o$ under the guidance of the weighted knowledge propagation graph $H_o$:

$$\tilde{\mathbf{z}}_o^i = \sum_{\mathbf{z}_o^j \in \mathcal{O}} H_o^{ij} \mathbf{z}_o^j, \quad (8)$$

where $\tilde{\mathbf{z}}_o^i$ and $\mathbf{z}_s^i$ have the same label. The augmented embeddings through *Cross-Domain Knowledge Propagation* raise the set $\tilde{\mathcal{D}}_o^f = \{\tilde{\mathbf{z}}_o^i | \mathbf{x}_s^i \in \mathcal{D}_s^f\}$ to augment $\mathcal{D}_s^f$.

*3) Cross-Domain Fulfillment Augmentation:* Moreover, besides the extremely imbalanced distributed source domain data, the different data distribution across source and target domains is another crucial challenge in imbalanced domain adaptation problems. Existing image generation strategies designed for few-shot problems, such as F2GAN [16], train a generator with images from the seen categories mapping a few conditional images to synthetic samples belonging to the same category. Then the trained model translates the images from unseen categories to diverse images with random interpolation coefficients. Such augmentation strategy does not take advantages of the discriminative knowledge in the annotated source domain data, and cannot manage the cross-domain distribution difference. Recent image recognition works reveal that the features deep in networks are usually linearized, and various directions in the feature space correspond to some specific semantic translations [54]. Such intriguing observation motivates the thoughts that translating one sample along specific feature direction resulting in new synthesized data with different semantics but still lying in the same class. Moreover, Xu *et al.* notice that in DA tasks, only samples from source and target domain alone are not sufficient to ensure domain-invariance at most part of latent space [55], which inspires us to generate synthesized data involving cross-domain information to fulfill the gap between source and target domain, as well as guarantee domain-invariance in a more continuous latent space.

However, due to the extremely imbalanced distribution in source domain and lacking of data from some specific categories causes directly translating the information across domains is vulnerable to negative transfer, especially when the available few-shot categories samples cannot represent the specific class distribution characteristics because they may lie far from the class center in the feature space. Thus aiming to implement a moderate augmentation strategy without severely misleading under the imbalance domain adaptation situation, we seek to generate synthesized samples through the feature level mix-up involving the augmented embeddings refined with knowledge propagation within the source domain and across source and target domains, i.e., $\tilde{\mathcal{D}}_s^f$ and $\tilde{\mathcal{D}}_o^f$.

Specifically, for each source domain sample $\mathbf{x}_s^i \in \mathcal{D}_s^f$ drawn from the minority set categories, two refined augmentation embeddings $\tilde{\mathbf{z}}_s^i \in \tilde{\mathcal{D}}_s^f$ and $\tilde{\mathbf{z}}_o^i \in \tilde{\mathcal{D}}_o^f$ are synthesized through the within-source and cross-domain structure knowledge propagation, respectively. To explore the internal information across domains, these two synthesized embeddings are linearly interpolated to fulfill the feature space across domains and produce mixup samples as:

$$\hat{\mathbf{z}}_m^i = (1 - \gamma)\tilde{\mathbf{z}}_s^i + \gamma\,\tilde{\mathbf{z}}_o^i, \tag{9}$$

where $\gamma \sim \text{Beta}(a, b)$ is to control the interpolation between the two embeddings ($\gamma \in [0, 1], a, b > 0$). Because $\tilde{\mathbf{z}}_s^i$ and $\tilde{\mathbf{z}}_o^i$ have identical class labels, the mixup samples are also assigned the same class label. For each source domain minority set sample $\mathbf{x}_s^i$, $k$ different mixup samples with label $\mathbf{y}_s^i$, the same label as $\mathbf{x}_s^i$, are generated with different randomly selected factor $\gamma$. The synthesized samples created by the cross-domain fulfillment augmentation constitute a new set denoted as $\hat{\mathcal{D}}_m^f = \{\hat{\mathbf{z}}_m^{i(k)} | \gamma^k \sim \text{Beta}(a, b)\}$, which is used to optimize the frameworks with corresponding class label.

It is noteworthy that our proposed *Cross-domain Fulfillment Augmentation* is different from DM-ADA proposed in [55]. Due to the extremely imbalanced distribution in the source domain caused by the lack of labeled samples from the minority set categories, directly combining samples across domains could produce fake samples leaning towards the majority set feature space severely. Moreover, the risk that the given few-shot samples lying far from the class center in the corresponding feature space will mislead the augmentation and domain alignment process. With our proposed strategy, the interpolated samples are produced based on the refined embeddings obtained with the within source and across source and target domain knowledge propagation under the guidance of the weighted graph, which can eliminate the undesired noise and distraction caused by outliers. Such a strategy can balance the contribution of the majority and minority set during the domain adaptation process.

*4) Hybrid Distinct Classifiers:* Recent works explored that dual classifiers structure can benefit the task specific decision boundaries and boost the domain adaptation performance [5], [19], [20]. However, in the imbalance domain adaptation problems, the extreme class-wise distribution imbalance make it impossible to train a promising classifier only based on given few-shot samples from the minority set. Even with the help of the knowledge propagation augmentation and cross-domain fulfillment strategies proposed in our work, the synthesized samples still cannot work as efficient as the real labeled data as the samples from the source domain majority set categories. Thus in our work, we propose a hybrid distinct classifiers mechanism to address this issue, which will evaluate the majority set and minority set categories from different perspectives separately. The hybrid distinct classifiers mechanism consists of $C_N(\cdot)$, a multi-layer neural network classifier, and $C_P(\cdot)$, a prototype classifier. The prototype classifier $C_P(\cdot)$ recognizes the target samples based on the similarity between each target sample to each category prototype, i.e., class center, instead of relying on the massive training data as $C_N(\cdot)$ does, which can eliminate the drawbacks caused by the insufficiency of the training data from the source domain minority set. On the contrary, for the majority set categories, the sufficient source domain training data can train a promising classifier for us to adapt to the target domain.

The probability prediction of $C_N(\cdot)$ and $C_P(\cdot)$ for each target sample $\mathbf{z}_t$ belonging to class $c$ are calculated as:

$$\hat{y}_{N,t}^c = \frac{e^{\psi^c(\mathbf{f}_t, \theta_{C_N})}}{\sum_{j=1}^{C} e^{\psi^j(\mathbf{f}_t, \theta_{C_N})}}, \quad \hat{y}_{P,t}^c = \frac{e^{\phi(\mathbf{f}_t, \mu_s^c)}}{\sum_{j=1}^{C} e^{\phi(\mathbf{f}_t, \mu_s^j)}}, \tag{10}$$

in which $\mathbf{f}_t$ is the output of the feature generator $F(\cdot)$, and $\theta_{C_N}$ denotes the trainable parameters in $C_N(\cdot)$, $\psi^c(\cdot, \cdot)$ is the $c$-th output of the last layer in $C_N(\cdot)$ before the softmax operation with input $\mathbf{f}_t$. $\phi(\cdot, \cdot)$ is defined to measure the similarity between the target sample to each task-specific prototype, and $\mu_s^c$ is the initialized source domain prototype of class $c$, $\mu^c = \frac{1}{n_s^c}\sum_{i=1}^{n_s^c} \mathbf{f}_{s}^{i,c}$. Since $C_P(\cdot)$ has no trainable parameters, we only apply supervision optimization to update the parameters in generator $F(\cdot)$ and classifier $C_N(\cdot)$ by minimizing the cross-entropy loss, which is defined as:

$$\mathcal{M}_s = -\frac{1}{\tilde{n}_s}\sum_{i=1}^{\tilde{n}_s}\sum_{c=1}^{C} \mathbf{1}_{[c=y_s^i]} \log \hat{y}_{N,s}^{i,c}, \tag{11}$$

where $\tilde{n}_s = n_s + \tilde{n}_s^f + \tilde{n}_o^f + \hat{n}_m^f$, is the total number of samples including all real source domain data ($n_s$) as well as synthesized samples through the within source *Embedding*

Propagation($\tilde{n}_s^f$), *Cross-domain Knowledge Propagation* ($\tilde{n}_o^f$), and *Cross-domain Fulfillment Augmentation*($\hat{n}_m^f$).

### D. Cross-Domain Prototypes Alignment (CPA)

So far, we have sufficient well-labeled source domain majority-set samples, few-shot minority-set real samples, together with the synthesized samples to make up the minority-set categories. In order to simultaneously solve the domain distribution disparity problem and augment the minority set data, we explore the Cross-Domain Prototype Alignment (CPA) strategy. Specifically, we accept the prototypes $\boldsymbol{\mu}_s^c$ as the dominant sector during the domain adaptation process and seek to align the source and target domain task-specific prototypes. Empirical Maximum Mean Discrepancy (MMD) [12], [25], [31] has been considered as a promising technique to align the source and target domain data domain-wise mean and class-wise mean. But the fact that source domain minority set categories with only few-shot labeled data available makes the MMD alignment not reliable anymore. Fortunately, we already have the synthesized samples making up the source domain minority-set feature distribution space, which will alleviate the dominance of the majority set over the minority set during the cross-domain alignment.

Specifically, for class $c$ in the source domain minority set label space $\mathcal{Q}^f$, the amended class prototype is calculated as:

$$\tilde{\boldsymbol{\mu}}_s^c = \frac{\sum\limits_{\mathbf{z}_s^i \in \mathcal{D}_s^{f(c)}} \mathbf{f}_s^i + \sum\limits_{\tilde{\mathbf{z}}_s^i \in \tilde{\mathcal{D}}_s^{f(c)}} \tilde{\mathbf{f}}_s^i + \sum\limits_{\tilde{\mathbf{z}}_o^i \in \tilde{\mathcal{D}}_o^{f(c)}} \tilde{\mathbf{f}}_o^i + \sum\limits_{\hat{\mathbf{z}}_m^i \in \hat{\mathcal{D}}_m^{f(c)}} \hat{\mathbf{f}}_m^i}{n_s^{f(c)} + \tilde{n}_s^{f(c)} + \tilde{n}_o^{f(c)} + \hat{n}_m^{f(c)}}, \quad (12)$$

where $\mathbf{f}_s^i/\tilde{\mathbf{f}}_s^i/\tilde{\mathbf{f}}_o^i/\hat{\mathbf{f}}_m^i$ are the output of the network $F(\cdot)$ with $\mathbf{z}_s^i/\tilde{\mathbf{z}}_s^i/\tilde{\mathbf{z}}_o^i/\hat{\mathbf{z}}_m^i$ as input, $\mathcal{D}_s^{f(c)}/\tilde{\mathcal{D}}_s^{f(c)}/\tilde{\mathcal{D}}_o^{f(c)}/\hat{\mathcal{D}}_m^{f(c)}$ are the subset of samples belonging to class $c$ drawn from $\mathcal{D}_s^f/\tilde{\mathcal{D}}_s^f/\tilde{\mathcal{D}}_o^f/\hat{\mathcal{D}}_m^f$, respectively, while $n_s^{f(c)}/\tilde{n}_s^{f(c)}/\tilde{n}_o^{f(c)}/\hat{n}_m^{f(c)}$ are the corresponding number of samples in each subset.

If all the source and target domain data are treated as a large batch during the augmentation process, global structure information will be propagated within and across domains. If so, $\tilde{n}_s^{f(c)} = \tilde{n}_o^{f(c)} = 1$ and $\hat{n}_m^{f(c)} = K$ in each training epoch. However, for large scale benchmarks, it is inefficient to deal with all data together and calculate the global knowledge propagation graph. Thus to handle large scale datasets and reduce the complexity, we can build *episodes* training strategy referring to few-shot learning tasks [52], [56], [57]. Each episode consists of $e_s$ source domain instances and $e_t$ instances from the target domain. For the source data in each episode, data from the majority set $\mathcal{Q}^m$ are randomly sampled without replacement ($p$ examples per class), and data belonging to the minority set $\mathcal{Q}^f$ can used multiple times in each training epoch due to the lack of data ($q$ examples per class), i.e., $e_s = |\mathcal{Q}^m| * p + |\mathcal{Q}^f| * q$. The computing complexity will be negligible with the episodes training strategy since the size of each episode is small [52]. The source domain prototypes are updated adaptively during training.

Based on the revised source domain class prototypes, the class-wise MMD can be calculated as:

$$\mathcal{M}_c = \frac{1}{C} \sum\nolimits_{c=1}^{C} \| \tilde{\boldsymbol{\mu}}_s^c - \boldsymbol{\mu}_t^c \|_2^2, \quad (13)$$

where $\tilde{\boldsymbol{\mu}}_s^c$ is the revised source domain class $c$ prototype, calculated on $\tilde{\mathcal{D}}_s^c = \{\mathcal{D}_s^{f(c)}, \tilde{\mathcal{D}}_s^{f(c)}, \tilde{\mathcal{D}}_o^{f(c)} \hat{\mathcal{D}}_o^{f(c)}\}$ when class $c$ belongs to the minority set categories. Due to the missing of the target domain samples labels, we accept the $C_P(\cdot)$ prediction $\hat{\mathbf{y}}_P^t$ as pseudo labels for target domain samples, and compute the average of the features predicted belonging to the same category as the corresponding prototype.

Beyond that, we extend to explicitly consider the data distribution among different classes across domains, and maximize the inter-class divergence as:

$$\mathcal{M}_d = \frac{1}{C} \frac{1}{C-1} \sum\nolimits_{c=1}^{C} \sum\nolimits_{\substack{c'=1 \\ c' \neq c}}^{C} \| \tilde{\boldsymbol{\mu}}_s^c - \boldsymbol{\mu}_t^{c'} \|_2^2, \quad (14)$$

where the inter-class divergence $\mathcal{M}_d$ evaluates the distances of all different class prototype pairs across domains.

Overall, our discriminative Cross-Domain Prototype Alignment is proposed to minimize the cross-domain intra-class prototypes distances, while maximizing the inter-class distances.

### E. Overall Objective and Optimization

To sum up, by exploring the source supervision over the real and augmented instances, as well as the discriminative cross-domain alignment, we have our overall objective function as:

$$\min_{F, C_N} \mathcal{M}_s + \lambda(\mathcal{M}_c - \mathcal{M}_d), \quad (15)$$

where $\lambda$ is a hyper-parameter to balance the contributions of different terms. We need to train the generator $F(\cdot)$ to map both source and target domain samples to a shared domain-invariant embedding feature space. It is noteworthy that the prototype classifier $C_P(\cdot)$ is free of parameters. Inspired by [5], [19], our training process consists of two steps:

*Step A:* We train the feature generator $F(\cdot)$ and neural networks classifier $C_N(\cdot)$ over the source supervision, including real and synthesized data. Keeping the performance on the source domain is crucial for obtaining discriminative whilst domain-invariant embedding features. Moreover, due to the lack of source domain minority-set samples, optimizing the whole model over the real samples as well as the synthesized samples will benefit the performance on the minority-set categories, avoiding the model overfitting to the majority-set classes. The optimization objective is listed as $\min_{F, C_N} \mathcal{M}_s$.

*Step B:* We freeze the parameters of classifier $C_N(\cdot)$ and update the generator $F(\cdot)$, which will map the source and target domain samples into a shared embedding feature space, where both source and target samples from the same class will be distributed close to each other, while far from samples belonging to other categories. The optimization objective is provided as $\min_F \mathcal{M}_s + \lambda(\mathcal{M}_c - \mathcal{M}_d)$.

## IV. EXPERIMENTS

We evaluate our proposed model on two domain adaptation visual benchmarks, and compare the evaluation performances with several state-of-the-art domain adaptation methods. Then we analyze the components of our framework in detail and explore the parameters sensitivity.

TABLE I

COMPARISONS OF RECOGNITION RATES (%) ON OFFICE-HOME DATASET (5-SHOT)

| Method | DAN | | | DM-ADA | | | MCD | | | SWD | | | SymNets | | | Ours | | |
|---|---|---|---|---|---|---|---|---|---|---|---|---|---|---|---|---|---|---|
| Acc | $A_f$ | $A_m$ | $A_o$ | $A_f$ | $A_m$ | $A_o$ | $A_f$ | $A_m$ | $A_o$ | $A_f$ | $A_m$ | $A_o$ | $A_f$ | $A_m$ | $A_o$ | $A_f$ | $A_m$ | $A_o$ |
| Ar→Cl | 22.1 | **57.1** | 43.6 | 5.13 | 20.2 | 14.4 | 14.5 | 32.7 | 25.7 | 15.3 | 47.3 | 35.0 | 13.9 | 39.6 | 30.3 | **47.7** | 52.9 | **50.9** |
| Ar→Pr | 30.5 | **71.7** | 55.1 | 2.35 | 28.2 | 17.8 | 26.6 | 49.9 | 40.5 | 20.1 | 67.1 | 48.2 | 36.5 | 64.4 | 55.4 | **71.3** | **74.0** | **72.9** |
| Ar→Rw | 42.0 | **77.5** | 62.7 | 4.31 | 37.1 | 23.5 | 35.1 | 51.3 | 44.6 | 30.9 | 70.8 | 54.2 | 40.8 | 71.8 | 58.2 | **72.0** | 77.2 | **75.0** |
| Cl→Ar | 15.7 | **60.6** | 40.4 | 6.52 | 20.3 | 14.1 | 17.8 | 28.3 | 23.6 | 11.8 | 46.2 | 30.8 | 9.60 | 58.7 | 37.2 | **57.3** | 60.6 | **59.1** |
| Cl→Pr | 25.7 | **72.0** | 52.7 | 0.00 | 29.3 | 17.5 | 6.50 | 67.0 | 42.7 | 2.20 | 38.5 | 23.9 | 12.3 | 67.1 | 46.6 | **65.1** | **74.8** | **70.9** |
| Cl→Rw | 18.4 | **73.2** | 51.1 | 0.00 | 28.1 | 16.4 | 15.2 | 12.8 | 13.8 | 4.10 | 34.3 | 21.7 | 20.4 | 66.1 | 48.4 | **69.3** | **74.7** | **72.5** |
| Pr→Ar | 22.0 | **57.0** | 41.3 | 1.01 | 23.8 | 13.6 | 27.4 | 14.0 | 20.0 | 28.2 | 39.0 | 34.2 | 18.1 | 51.0 | 37.8 | **60.9** | 58.8 | **59.8** |
| Pr→Cl | 20.2 | **52.3** | 40.0 | 5.85 | 27.1 | 19.0 | 4.50 | 35.1 | 23.4 | 3.90 | 36.5 | 17.8 | 7.30 | 40.9 | 26.5 | **46.0** | 51.4 | **49.3** |
| Pr→Rw | 42.6 | **77.8** | 36.1 | 3.42 | 43.4 | 26.8 | 36.7 | 54.1 | 46.9 | 13.4 | 74.6 | 49.2 | 38.4 | 70.9 | 58.6 | **72.6** | **80.5** | **77.2** |
| Rw→Ar | 26.5 | **68.2** | 49.5 | 6.52 | 41.2 | 25.6 | 27.0 | 44.0 | 36.4 | 25.1 | 60.8 | 44.8 | 27.4 | 65.4 | 50.0 | **63.0** | 66.3 | **64.9** |
| Rw→Cl | 25.0 | **60.9** | 47.1 | 2.05 | 31.0 | 19.9 | 4.10 | 35.9 | 23.7 | 3.00 | 50.5 | 32.3 | 14.9 | 46.8 | 33.3 | **49.9** | 54.9 | **53.0** |
| Rw→Pr | 38.8 | **84.9** | 66.3 | 1.29 | 55.6 | 33.8 | 32.2 | 59.0 | 48.2 | 7.60 | 50.9 | 33.5 | 30.7 | 81.0 | 60.8 | **76.8** | 84.0 | **81.1** |
| Avg. | 30.9 | 64.3 | 51.1 | 3.20 | 32.1 | 20.2 | 20.6 | 40.3 | 32.4 | 13.8 | 50.5 | 35.5 | 22.5 | 60.3 | 45.3 | **62.7** | **67.5** | **65.6** |

## A. Datasets and Experimental Setting

*Office-31 [58]:* Is a standard benchmark dataset for visual domain adaptation. It contains 4,110 images from 31 categories, shared by three domains: Amazon (A), Webcam (W) and DSLR (D). We evaluate our methods following the common protocol on all 6 adaptation tasks.

*Office-Home [59]:* Is a more various yet challenging benchmark, consisting of 15,500 images from 65 classes of daily objects in office and home, shared by four extremely distinct domains: Artistic (Ar), Clip Art (Cl), Product (Pr), and Real-World (Rw). The model is evaluated on all 12 tasks.

*Implementation Details:* We implement our model based on PyTorch. ImageNet pre-trained ResNet-50 [47] without the last fully-connected layer is accepted as $E(\cdot)$ to obtain the embeddings $\mathbf{z}_{s/t}$. $F(\cdot)$ is a two-layer fully-connected neural networks with hidden layer output dimension as 1,024 and ReLU activation. The output of $F(\cdot)$ is the domain invariant features $\mathbf{f}_{s/t}$ with dimension 512. $C_N(\cdot)$ is a two-layer fully-connected neural networks classifier with hidden layer dimension as 512. Cosine similarity is accepted as the measurement function $\phi(\cdot)$ in the prototype classifier $C_P(\cdot)$. We take the embedding features mean of the source domain samples belonging to each category as the initialized prototype $\boldsymbol{\mu}_s^c$ which will be used by $C_P(\cdot)$ for classification, while for the Cross-Domain Prototype Alignment, we take the synthesized samples into account to update the prototype $\boldsymbol{\mu}_s^c \rightarrow \tilde{\boldsymbol{\mu}}_s^c$. All trainable parameters are optimized by Adam optimizer with learning rate as 0.001 for both Office-31 and Office-Home dataset. $F(\cdot)$ and $C_N(\cdot)$ are initialized and pre-trained on the source domain with learning rate as 0.0001 for 2,000 epochs, while $E(\cdot)$ is fixed. $\lambda$ is fixed as 0.1 for Office-Home, while 0.01 for Office-31, $\gamma \sim \text{Beta}(2, 2)$, and $k$ is fixed as 5, which will be discussed in the ablation study section. To decide the optimal values of hyper-parameters, we follow [31] to construct a validation set with labeled source and unlabeled target data, then a binary domain classifier is trained to distinguish the source and target data on the validation set. The source classifier error and domain classifier error are jointly assessed to determine the parameters for each task.

We focus on two challenging extremely-imbalanced domain adaptation tasks as 1-shot and 5-shot in our experiments. Specifically, for each domain transfer task, we randomly select 1 or 5 samples of each source domain minority-set category, together with all the rest labeled majority set source domain data as well as the unlabeled target domain samples for training. The first 10 and the first 25 alphabetically classes are treated as the minority-set in the Office-31 and Office-Home dataset, respectively, the rest classes constitute the majority-set. We randomly run the experiments 3 times and report the average results of the $30^{th}$ epoch. For each case, we report the results of $C_N(\cdot)$ on the target domain majority-set categories, while for the minority-set categories, we show the results of $C_P(\cdot)$, and the overall average performance is also based on these two results, as illustrated in the *Hybrid Distinct Classifiers*. We will discuss the different specialities of $C_N(\cdot)$ and $C_P(\cdot)$ in the ablation study section. All baselines are implemented with the official codes with hyper-parameters tuning as instructed by the original papers. For all experiments results, we mark the best performance on the minority-set as **blue**, the best performance on the majority-set as **red**, and the best overall performance as **bold**.

## B. Results and Comparisons

The classification results on Office-Home and Office-31 under 5-shot and 1-shot settings are reported in Tables I, II, III, and IV, respectively. $A_f$ means minority-set accuracy, $A_m$ represents majority-set accuracy, and $A_o$ denotes the overall accuracy on the whole target domain.

From the results, it is obvious that our method significantly outperforms all the comparisons on both two benchmarks under 4 challenging settings in terms of the overall accuracy. Especially for the performance of the minority-set, our model achieves promising results while keeping reliable performance on the majority-set, which emphasizes the robustness of our model to manage the extremely imbalanced distribution challenges. On the contrary, conventional UDA solutions, e.g., DM-ADA [55] and SymNets [20], suffer from the source distribution imbalance problem, and fail to perform well in the minority-set categories, due to no consideration of imbalance distribution. For the Office-31 dataset 5-shot tasks, our model achieves 93.0% average accuracy on the minority-set, which beats DAN over 26%, and maintains 87.3% on the majority-set, which is higher than SymNets. Furthermore, our model

| Method | DAN | | | DM-ADA | | | MCD | | | SWD | | | SymNets | | | Ours | | |
|---|---|---|---|---|---|---|---|---|---|---|---|---|---|---|---|---|---|---|
| Acc | $A_f$ | $A_m$ | $A_o$ | $A_f$ | $A_m$ | $A_o$ | $A_f$ | $A_m$ | $A_o$ | $A_f$ | $A_m$ | $A_o$ | $A_f$ | $A_m$ | $A_o$ | $A_f$ | $A_m$ | $A_o$ |
| Ar→Cl | 0.30 | 56.7 | 35.0 | 2.81 | 23.1 | 15.3 | 14.3 | 48.3 | 35.3 | 15.3 | 45.7 | 34.0 | 0.40 | 43.5 | 27.2 | 29.4 | 54.8 | 45.1 |
| Ar→Pr | 4.00 | 72.0 | 44.7 | 1.18 | 28.3 | 17.4 | 23.0 | 65.9 | 48.5 | 23.4 | 66.6 | 49.2 | 0.00 | 69.4 | 41.5 | 46.2 | 76.3 | 64.2 |
| Ar→Rw | 4.20 | 78.0 | 47.3 | 0.00 | 39.2 | 22.9 | 32.4 | 70.3 | 54.5 | 31.4 | 71.3 | 54.7 | 0.00 | 75.0 | 43.7 | 53.9 | 78.7 | 68.4 |
| Cl→Ar | 0.00 | 60.6 | 33.4 | 0.00 | 18.6 | 10.3 | 16.8 | 33.8 | 26.2 | 12.1 | 47.0 | 31.3 | 0.00 | 60.1 | 32.9 | 41.2 | 60.5 | 51.8 |
| Cl→Pr | 0.00 | 73.5 | 44.0 | 2.30 | 28.8 | 18.1 | 4.90 | 21.9 | 15.1 | 5.40 | 68.8 | 43.3 | 0.00 | 74.3 | 44.2 | 43.8 | 76.0 | 63.0 |
| Cl→Rw | 0.00 | 71.5 | 41.9 | 1.77 | 25.0 | 15.4 | 17.7 | 42.2 | 32.0 | 3.40 | 65.5 | 39.7 | 0.00 | 71.0 | 41.7 | 50.7 | 76.0 | 65.5 |
| Pr→Ar | 0.00 | 57.6 | 31.8 | 0.00 | 26.2 | 14.5 | 25.2 | 50.2 | 39.0 | 25.0 | 49.1 | 38.3 | 0.00 | 56.5 | 31.8 | 47.6 | 60.1 | 54.5 |
| Pr→Cl | 0.80 | 53.2 | 33.1 | 0.00 | 29.3 | 18.0 | 4.30 | 47.0 | 30.6 | 4.10 | 46.3 | 30.1 | 0.00 | 45.0 | 27.0 | 34.8 | 52.3 | 45.6 |
| Pr→Rw | 7.50 | 77.3 | 48.3 | 3.64 | 40.8 | 25.3 | 36.5 | 74.8 | 57.7 | 14.9 | 74.7 | 49.5 | 0.20 | 74.4 | 43.6 | 63.8 | 80.5 | 73.6 |
| Rw→Ar | 0.00 | 68.5 | 37.8 | 2.02 | 37.3 | 21.5 | 26.7 | 39.0 | 33.5 | 25.9 | 61.6 | 45.6 | 71.5 | 39.3 | 56.9 | 56.9 | 66.7 | 62.3 |
| Rw→Cl | 2.70 | 61.5 | 39.0 | 0.00 | 31.7 | 19.5 | 4.10 | 19.2 | 13.4 | 3.10 | 11.3 | 8.20 | 0.00 | 54.1 | 33.1 | 40.2 | 57.3 | 50.7 |
| Rw→Pr | 1.90 | 84.6 | 51.3 | 1.74 | 56.3 | 34.4 | 36.0 | 80.6 | 62.7 | 6.3 | 80.7 | 50.8 | 0.00 | 84.4 | 41.7 | 59.8 | 84.6 | 74.6 |
| Avg. | 1.80 | 67.9 | 40.6 | 1.29 | 32.0 | 19.4 | 20.2 | 49.2 | 37.4 | 14.2 | 57.3 | 39.6 | 0.10 | 64.9 | 37.3 | 47.4 | 68.7 | 59.9 |

| Method | DAN | | | DM-ADA | | | MCD | | | SWD | | | SymNets | | | Ours | | |
|---|---|---|---|---|---|---|---|---|---|---|---|---|---|---|---|---|---|---|
| Acc | $A_f$ | $A_m$ | $A_o$ | $A_f$ | $A_m$ | $A_o$ | $A_f$ | $A_m$ | $A_o$ | $A_f$ | $A_m$ | $A_o$ | $A_f$ | $A_m$ | $A_o$ | $A_f$ | $A_m$ | $A_o$ |
| A→W | 41.2 | 77.1 | 67.1 | 9.65 | 51.9 | 38.8 | 56.6 | 6.30 | 21.9 | 44.3 | 7.30 | 18.7 | 17.0 | 92.1 | 69.9 | 97.3 | 90.3 | 92.4 |
| D→W | 99.6 | 96.3 | 97.2 | 3.70 | 73.9 | 52.2 | 63.4 | 57.0 | 61.1 | 54.9 | 22.5 | 32.5 | 76.2 | 98.6 | 92.1 | 99.7 | 98.6 | 99.0 |
| W→D | 92.9 | 99.7 | 97.6 | 0.00 | 85.7 | 60.3 | 31.2 | 37.5 | 35.6 | 51.3 | 79.1 | 70.9 | 79.9 | 100 | 93.8 | 100 | 99.3 | 99.5 |
| A→D | 47.4 | 85.5 | 73.7 | 6.73 | 49.6 | 36.9 | 56.5 | 14.5 | 26.9 | 58.4 | 72.4 | 68.3 | 13.6 | 88.7 | 65.5 | 97.4 | 89.8 | 92.2 |
| D→A | 64.6 | 62.1 | 62.8 | 0.52 | 35.9 | 25.6 | 63.6 | 43.0 | 49.0 | 53.5 | 64.3 | 61.1 | 53.8 | 71.4 | 65.9 | 81.3 | 73.6 | 75.9 |
| W→A | 53.3 | 63.7 | 60.7 | 3.68 | 43.1 | 31.5 | 68.2 | 24.9 | 37.6 | 58.5 | 42.7 | 47.3 | 49.8 | 69.7 | 63.1 | 82.2 | 72.2 | 75.1 |
| Avg. | 66.5 | 80.7 | 76.5 | 4.05 | 56.7 | 40.9 | 56.6 | 31.0 | 38.7 | 53.5 | 48.1 | 49.8 | 48.4 | 86.8 | 75.1 | 93.0 | 87.3 | 89.0 |

| Method | DAN | | | DM-ADA | | | MCD | | | SWD | | | SymNets | | | Ours | | |
|---|---|---|---|---|---|---|---|---|---|---|---|---|---|---|---|---|---|---|
| Acc | $A_f$ | $A_m$ | $A_o$ | $A_f$ | $A_m$ | $A_o$ | $A_f$ | $A_m$ | $A_o$ | $A_f$ | $A_m$ | $A_o$ | $A_f$ | $A_m$ | $A_o$ | $A_f$ | $A_m$ | $A_o$ |
| A→W | 0.00 | 76.3 | 53.7 | 4.11 | 50.3 | 36.0 | 55.7 | 4.30 | 20.2 | 41.3 | 72.3 | 62.7 | 0.00 | 91.1 | 64.2 | 85.7 | 88.2 | 87.5 |
| D→W | 48.9 | 96.2 | 82.2 | 0.00 | 71.8 | 49.6 | 63.0 | 24.5 | 36.4 | 45.1 | 96.8 | 80.8 | 0.00 | 98.4 | 69.7 | 97.0 | 98.8 | 98.2 |
| W→D | 37.7 | 99.7 | 80.5 | 0.00 | 76.3 | 53.7 | 35.7 | 99.4 | 80.6 | 38.3 | 99.7 | 81.6 | 0.00 | 99.7 | 68.8 | 97.2 | 99.5 | 98.8 |
| A→D | 0.00 | 83.7 | 57.8 | 0.00 | 49.3 | 34.7 | 57.1 | 14.5 | 27.1 | 52.6 | 75.9 | 69.0 | 0.00 | 93.6 | 64.7 | 85.1 | 89.2 | 88.0 |
| D→A | 26.8 | 63.4 | 52.7 | 0.00 | 48.8 | 28.9 | 67.4 | 18.0 | 32.4 | 53.0 | 46.3 | 48.3 | 0.00 | 68.3 | 47.9 | 79.7 | 74.1 | 75.8 |
| W→A | 12.7 | 64.5 | 49.4 | 0.00 | 38.4 | 27.1 | 65.4 | 32.4 | 42.1 | 56.8 | 61.9 | 60.4 | 0.00 | 67.5 | 47.5 | 80.4 | 72.1 | 74.5 |
| Avg. | 21.0 | 80.7 | 62.7 | 0.68 | 54.5 | 38.3 | 57.4 | 32.2 | 39.8 | 47.9 | 74.5 | 67.1 | 0.15 | 86.4 | 60.5 | 87.5 | 87.0 | 87.1 |

gets promising overall performance on the Office-31 dataset as 89.0%, which is even comparable to the state-of-the-art performance 88.4% achieved by SymNets, which include more source domain minority set labeled samples for training [20]. Moreover, on another extremely challenging case when only 1 sample is available for each category in the minority-set, our model still gets reasonable results on the minority-set as well as stable performance on the majority-set. It highly affirms the effectiveness and robustness of our method dealing with domain adaptation problem in which the source domain data is extremely imbalanced and insufficient for training. From the results reported in Tables I - IV, we observe that the classification accuracy of proposed methods is much higher than the other compared baselines in most cases. This justifies the efficacy of the Cross-Domain Augmentation and the Cross-Domain Prototype Alignment in dealing with imbalanced domain adaptation challenges.

### C. Comparison With Imbalanced DA Solution

To demonstrate the effectiveness of the proposed TFKT, we show more results of source-only hybrid classifiers and Weighted Maximum Mean Discrepancy (WDAN) [27]. Source-only hybrid classifiers consider $C_N(\cdot)$ on the target domain majority-set categories, while the results of $C_P(\cdot)$ on the minority-set. WDAN manages the domain adaptation with data distribution imbalance issues through reweighing the importance of each source sample during the domain alignment process. We re-implement WDAN with ResNet-50 [47] as the backbone, as the ResNets are the preferred base networks contemporaneously.[1]

From the results in Tables V & VI, we observe that our TFKT beats all compared baselines in most cases and achieves the best average results. WDAN obtains good performance in the original imbalance situation claimed in [27], but it cannot handle the extreme situations claimed in our paper when only 1 or 5 samples per-class are available for training. In addition, we notice the Source-only results also surpass some conventional domain adaptation solutions, especially in the minority set, emphasizing the contribution and benefits of involving

---

[1] The original WDAN is implemented with LeNet [60], AlexNet [1], GoogleLeNet [61], and VGG16 [2] as the backbone.

TABLE V

COMPARISONS OF RECOGNITION RATES (%) ON OFFICE-31 DATASET (1-SHOT)

| Shot | 1 - shot | | | | | | | | | 5 - shot | | | | | | | | |
|---|---|---|---|---|---|---|---|---|---|---|---|---|---|---|---|---|---|---|
| Method | Source-only | | | WDAN | | | Ours | | | Source-only | | | WDAN | | | Ours | | |
| Acc | $A_f$ | $A_m$ | $A_o$ | $A_f$ | $A_m$ | $A_o$ | $A_f$ | $A_m$ | $A_o$ | $A_f$ | $A_m$ | $A_o$ | $A_f$ | $A_m$ | $A_o$ | $A_f$ | $A_m$ | $A_o$ |
| A→W | 68.51 | 78.04 | 75.09 | 13.19 | 81.07 | 61.01 | **85.7** | **88.2** | **87.5** | 73.19 | 77.68 | 76.29 | | | | **97.3** | **90.3** | **92.4** |
| D→W | 91.49 | 97.14 | 95.39 | 52.34 | 97.32 | 84.03 | **97.0** | **98.8** | **98.2** | 93.19 | 97.32 | 96.04 | 87.23 | 98.04 | 94.84 | **99.7** | **98.6** | **99.0** |
| W→D | 95.45 | 99.71 | 98.45 | 46.10 | | 83.33 | **97.2** | **99.5** | **98.8** | 96.75 | 99.71 | 98.84 | 89.61 | 99.42 | 96.39 | **100.0** | **99.3** | **99.5** |
| A→D | 77.92 | 79.65 | 79.14 | 18.83 | 76.74 | 58.84 | **85.1** | **89.2** | **88.0** | 79.22 | 76.54 | 77.33 | 51.30 | 82.85 | 73.09 | **97.4** | **89.8** | **92.2** |
| D→A | 65.66 | 66.63 | 66.53 | 27.31 | 71.15 | 58.32 | **79.7** | **74.1** | **75.8** | 67.60 | 66.73 | 66.80 | 60.92 | 70.65 | 67.80 | **81.3** | **73.6** | **75.9** |
| W→A | 66.50 | 63.92 | 64.67 | 11.29 | 66.68 | 50.51 | **80.4** | **72.1** | **74.5** | 69.66 | 62.82 | 64.82 | | 65.73 | 64.79 | **82.2** | **72.2** | **75.1** |
| Avg. | 77.59 | 80.85 | 79.85 | 28.18 | 82.16 | 66.01 | **87.5** | **87.0** | **87.1** | 79.94 | 80.13 | 80.05 | 65.26 | 82.90 | 77.60 | **93.0** | **87.3** | **89.0** |

TABLE VI

COMPARISONS OF RECOGNITION RATES (%) ON OFFICE-HOME DATASET (1-SHOT)

| Shot | 1 - shot | | | | | | | | | 5 - shot | | | | | | | | |
|---|---|---|---|---|---|---|---|---|---|---|---|---|---|---|---|---|---|---|
| Method | Source-only | | | WDAN | | | Ours | | | Source-only | | | WDAN | | | Ours | | |
| Acc | $A_f$ | $A_m$ | $A_o$ | $A_f$ | $A_m$ | $A_o$ | $A_f$ | $A_m$ | $A_o$ | $A_f$ | $A_m$ | $A_o$ | $A_f$ | $A_m$ | $A_o$ | $A_f$ | $A_m$ | $A_o$ |
| Ar→Cl | 26.39 | 43.79 | 37.11 | 0.00 | 49.52 | 30.52 | **29.4** | **54.8** | **45.1** | 30.09 | 47.40 | 40.76 | 13.49 | 47.96 | 34.73 | **47.7** | **52.9** | **50.9** |
| Ar→Pr | 41.29 | 62.36 | 53.89 | 0.06 | 70.80 | 42.35 | **46.2** | **76.3** | **64.2** | 41.79 | 68.09 | 57.51 | 19.27 | 70.80 | 50.08 | **71.3** | **74.0** | **72.9** |
| Ar→Rw | 49.25 | 72.82 | 63.02 | 0.09 | 75.96 | 44.39 | **53.9** | **78.7** | **68.4** | 41.79 | 71.21 | 58.98 | 25.40 | 76.04 | 54.99 | **72.0** | **77.2** | **75.0** |
| Cl→Ar | 32.32 | 51.57 | 42.93 | 0.00 | 56.13 | 30.94 | **41.2** | **60.5** | **51.8** | 39.03 | 50.90 | 45.57 | 7.35 | 55.53 | 33.91 | **57.3** | **60.6** | **59.1** |
| Cl→Pr | 34.90 | 68.69 | 55.10 | 0.00 | 73.47 | 43.93 | **43.8** | **76.0** | **63.0** | 40.11 | 69.37 | 57.60 | 0.34 | 73.32 | 43.97 | **65.1** | **74.8** | **70.9** |
| Cl→Rw | 40.03 | 67.36 | 56.00 | 0.00 | 72.03 | 42.09 | **50.7** | **76.0** | **65.9** | 47.05 | 67.75 | 59.15 | 4.64 | 72.27 | 44.16 | **69.3** | **74.7** | **72.5** |
| Pr→Ar | 44.72 | 53.89 | 49.78 | 0.00 | 55.31 | 30.49 | **47.6** | **60.1** | **54.5** | 49.49 | 52.39 | 51.09 | 0.18 | 56.58 | 31.27 | **60.9** | **58.8** | **59.8** |
| Pr→Cl | 27.94 | 48.51 | 40.62 | 0.00 | 50.07 | 30.86 | **34.8** | **52.3** | **45.6** | 31.58 | 47.11 | 41.40 | 0.00 | 49.48 | 30.49 | **46.0** | **51.4** | **49.3** |
| Pr→Rw | 60.57 | 74.86 | 68.92 | 0.00 | 77.61 | 45.35 | **63.8** | **80.5** | **73.6** | 62.62 | 75.18 | 69.96 | 0.00 | 76.75 | 44.85 | **72.6** | **80.5** | **77.5** |
| Rw→Ar | 44.08 | 64.65 | 55.42 | 0.00 | 65.47 | 36.09 | **56.9** | **66.7** | **62.3** | 50.60 | 65.40 | 58.76 | 5.05 | 65.70 | 38.48 | **63.0** | **66.3** | **64.9** |
| Rw→Cl | 30.27 | 51.34 | 43.25 | 0.00 | 53.57 | 33.01 | **40.2** | **57.3** | **50.7** | 33.55 | 52.01 | 44.93 | 2.93 | 52.04 | 33.20 | **49.9** | **54.9** | **53.0** |
| Rw→Pr | 55.74 | 82.10 | 71.50 | 0.00 | 83.65 | 50.01 | **59.8** | **84.6** | **74.6** | 62.35 | 82.18 | 74.21 | 2.30 | 84.14 | 51.20 | **76.8** | **84.0** | **81.1** |
| Avg. | 40.63 | 61.83 | 51.13 | 0.01 | 65.30 | 38.34 | **47.4** | **68.7** | **59.9** | 44.17 | 62.45 | 54.09 | 6.75 | 65.05 | 40.04 | **62.7** | **67.5** | **65.6** |

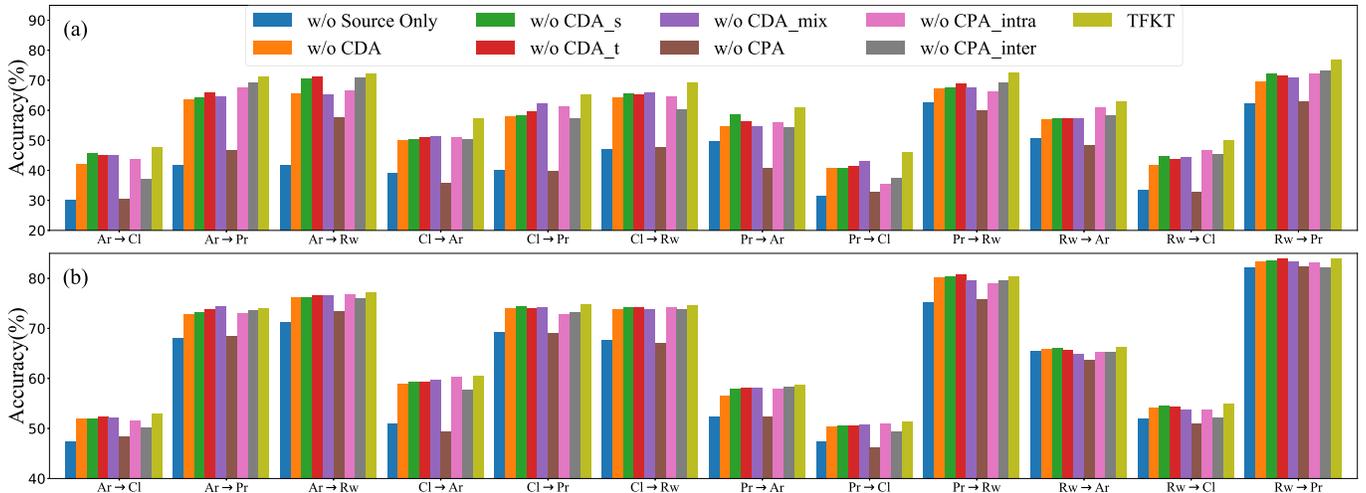

Fig. 4. Contribution of the Cross-Domain Prototype Alignment (CPA) and the Cross-domain Augmentation (CDA) strategy on Office-Home 25-way 5-shot tasks (a) $C_P(\cdot)$ performance on minority-set and (b) $C_N(\cdot)$ performance on majority-set.

the prototypical classifier $C_P(\cdot)$. The proposed *Hybrid Distinct Classifiers* framework can significantly counteract the negative effect caused by the training data insufficiency.

### D. Ablation Analysis

First of all, we evaluate the contribution of the Cross-domain Prototypes Alignment (CPA) and the Cross-domain Augmentation (CDA) to our model by removing one of them and keep all other architectures and training strategy. In Fig. 4, we remove CPA, or the CDA strategy, or both of them and show the results on Office-Home as "w/o CPA", "w/o CDA", and "Source Only", respectively. Besides, the source domain minority set data is augmented by CDA to three kinds of synthetic data $\tilde{\mathcal{D}}_s^f$, $\tilde{\mathcal{D}}_o^f$, and $\tilde{\mathcal{D}}_m^f$, through *Data Augmentation through Embedding Propagation*, *Cross-domain Knowledge Propagation*, and *Cross-domain Fulfillment Augmentation*, respectively. By removing each one kind of synthetic data while keeping others, the results are reported as "w/o CDA_s", "w/o CDA_t", "w/o CDA_mix", respectively. Moreover, CPA consists of two loss terms, minimizing class-wise MMD ($\mathcal{M}_c$) and maximizing inter-class divergence ($\mathcal{M}_d$). We remove each one term and the results are denoted as "w/o CPA_intra" and "w/o CPA_inter", respectively. Fig. 4 (a) claims the $C_P(\cdot)$ performance on the target domain minority-set classes, and (b) shows the $C_N(\cdot)$ performance on the majority-set. From the results, we observe that both CPA and CDA strategies benefit the fair cross-domain learning especially on the recognition performance on the minority set. It is reasonable that $C_N(\cdot)$ performance on the majority-set is not promoted significantly by CDA, because CDA only augments the minority-set categories. But the CPA strategy boosts the $C_P(\cdot)$ performance on the minority-set categories impressively.

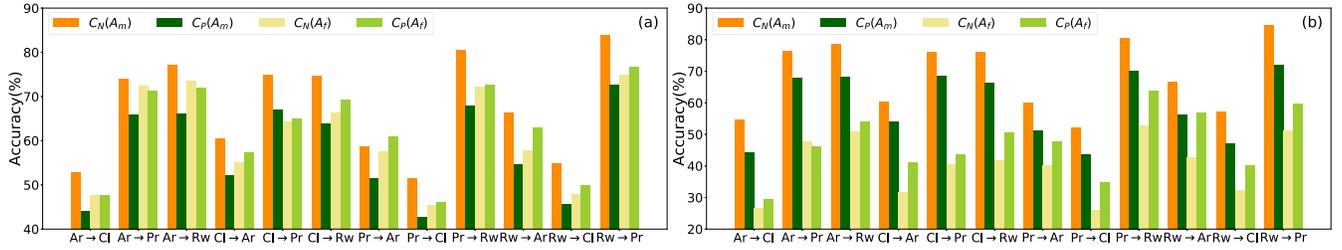

Fig. 5. $C_N(\cdot)$ and $C_P(\cdot)$ performance comparison on Office-Home majority- and minority-set (a) 5-shot, (b) 1-shot.

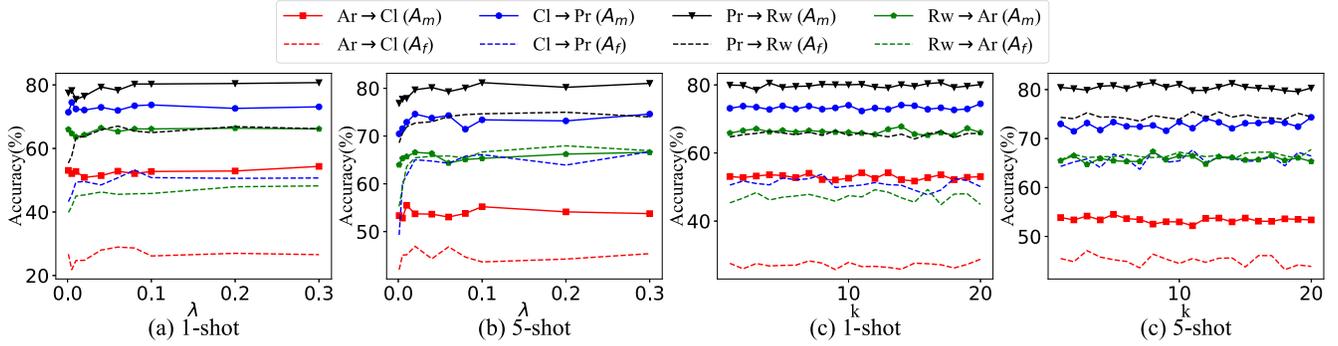

Fig. 6. Parameters sensitivity analysis of $\lambda$ and $k$ on the Office-Home selected tasks (a) (c) 1-shot, (b) (d) 5-shot.

Secondly, we compare the different classification specialities of $C_N(\cdot)$ and $C_P(\cdot)$ on different subsets. In Fig. 5, we show $C_N(\cdot)$ and $C_P(\cdot)$ recognition rate of the target domain majority-set and minority-set categories on the Office-Home 25-way 5-shot tasks. The main difference between the roles of neural network classifier $C_N(\cdot)$ and prototypical classifier $C_P(\cdot)$ is their classification ability on the minority-set categories. The insufficiency of training data from the majority-set classes makes the trained neural network classifier dominated by the majority set and fail on the minority set. On the contrary, the prototype classifier is based on the estimated prototype from given samples per category, which are decided by the quality of available data instead of the number of samples available. From the results, we notice that for categories with sufficient well-labeled source samples for training in the majority-set, $C_N(\cdot)$ always obtains better performance than the prototype classifier $C_P(\cdot)$, e.g., Pr→Cl. However, for those classes lacking training samples in the minority-set, $C_P(\cdot)$ can handle it much better and achieve promising performance in most cases, e.g., Pr→Cl and Rw→Cl. The generated samples contribute to refining the prototypes during training and benefit the classification performance of $C_P(\cdot)$. From the results, we notice that the improvement of $C_P(\cdot)$ compared to $C_N(\cdot)$ for the minority set is more significant on 1-shot setting than 5-shot. So the fewer source domain minority-set data available for training, choosing $C_P(\cdot)$ to recognize the minority set is more reasonable and superior.

Thirdly, we analyze the parameter sensitivity of our model. Four hard-to-transfer tasks of Office-Home dataset are used for evaluation. The results are reported in Fig. 6. We can see that transfer performance is not sensitive to the variance of hyper-parameter $\lambda$ from 0.1 to 0.3, in both 5-shot and 1-shot settings, which demonstrate the importance of the *Cross-domain Prototype Alignment*. Moreover, we change the number $k$ of generated fake samples in each class from 1 to 20. Fig. 6(c) and (d) show that the results are not sensitive to the number of fake samples generated by the *Cross-domain Fulfillment Augmentation* after $k = 5$.

Finally, we evaluate the quality of the generated synthesized samples belonging to the minority-set categories by drawing the t-SNE embeddings of all the real source and target samples, together with the generated fake samples. The results are visualized in Fig. 7. The red star points are the 1-shot set samples and the blue dots are the source domain majority set samples, gray dots are the target domain data. The augmented fake data are represented as yellow dots. (a) shows the output embeddings of network $E(\cdot)$, and (b) shows the output features of $F(\cdot)$. It is obvious that the generated samples are very similar to the available source domain minority-set samples, and the comparison between the chaos in (a) and organized data distribution in (b) demonstrate the effectiveness of the proposed cross-domain augmentation and prototypes alignment strategies.

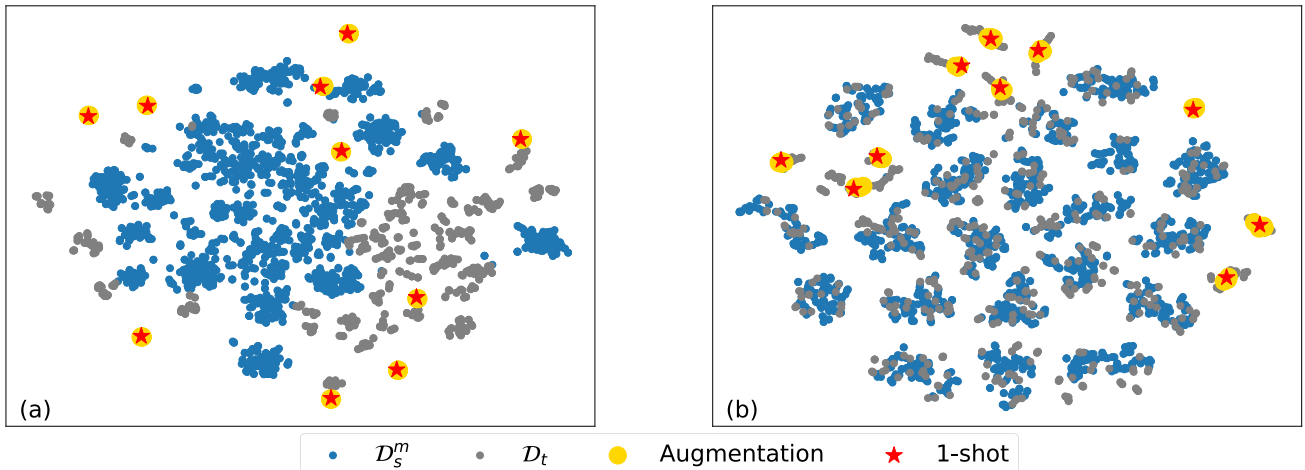

Fig. 7. Visualization of embedding features of the Office-31 dataset A→W 10-way 1-shot task, including real and augmented fake samples. (a) Embeddings output from $E(\cdot)$ (b) Features output from $F(\cdot)$.

## V. CONCLUSION

In this paper, we addressed the problem of fairness challenge in extremely imbalanced cross-domain learning by proposing a Towards Fair Knowledge Transfer (TFKT) model. Specifically, cross-domain feature augmentation strategy was proposed to augment the source domain minority-set categories by within-source and cross-domain embedding knowledge propagation and cross-domain fulfillment augmentation. Moreover, cross-domain prototypes alignment was further explored to seek a hybrid classifier framework towards fair knowledge transfer recognizing the target domain data. Extensive experiments on different tasks demonstrated the predominant performance of our model dealing with minority-set in domain adaptation.

## REFERENCES

[1] A. Krizhevsky, I. Sutskever, and G. E. Hinton, "Imagenet classification with deep convolutional neural networks," in *Proc. Adv. Neural Inf. Process. Syst.*, 2012, pp. 1097–1105.

[2] K. Simonyan and A. Zisserman, "Very deep convolutional networks for large-scale image recognition," 2014, *arXiv:1409.1556*. [Online]. Available: http://arxiv.org/abs/1409.1556

[3] M. Du, A. C. Sankaranarayanan, and R. Chellappa, "Robust face recognition from multi-view videos," *IEEE Trans. Image Process.*, vol. 23, no. 3, pp. 1105–1117, Mar. 2014.

[4] L. Wang, B. Sun, J. Robinson, T. Jing, and Y. Fu, "EV-action: Electromyography-vision multi-modal action dataset," in *Proc. 15th IEEE Int. Conf. Autom. Face Gesture Recognit.*, Nov. 2020, pp. 160–167.

[5] K. Saito, K. Watanabe, Y. Ushiku, and T. Harada, "Maximum classifier discrepancy for unsupervised domain adaptation," in *Proc. IEEE Conf. Comput. Vis. Pattern Recognit.*, Jun. 2018, pp. 3723–3732.

[6] S. Roy, A. Siarohin, E. Sangineto, S. R. Bulo, N. Sebe, and E. Ricci, "Unsupervised domain adaptation using feature-whitening and consensus loss," in *Proc. IEEE/CVF Conf. Comput. Vis. Pattern Recognit. (CVPR)*, Jun. 2019, pp. 9471–9480.

[7] Y. Pan, T. Yao, Y. Li, Y. Wang, C.-W. Ngo, and T. Mei, "Transferrable prototypical networks for unsupervised domain adaptation," in *Proc. IEEE/CVF Conf. Comput. Vis. Pattern Recognit. (CVPR)*, Jun. 2019, pp. 2239–2247.

[8] H. Feng, M. Chen, J. Hu, D. Shen, H. Liu, and D. Cai, "Complementary pseudo labels for unsupervised domain adaptation on person re-identification," *IEEE Trans. Image Process.*, vol. 30, pp. 2898–2907, 2021.

[9] C.-A. Hou, Y.-H. H. Tsai, Y.-R. Yeh, and Y.-C. F. Wang, "Unsupervised domain adaptation with label and structural consistency," *IEEE Trans. Image Process.*, vol. 25, no. 12, pp. 5552–5562, Dec. 2016.

[10] W. Zhang, X. Li, H. Ma, Z. Luo, and X. Li, "Open-set domain adaptation in machinery fault diagnostics using instance-level weighted adversarial learning," *IEEE Trans. Ind. Informat.*, vol. 17, no. 11, pp. 7445–7455, Nov. 2021.

[11] K. M. Borgwardt, A. Gretton, M. J. Rasch, H.-P. Kriegel, B. Scholkopf, and A. J. Smola, "Integrating structured biological data by kernel maximum mean discrepancy," *Bioinformatics*, vol. 22, no. 14, pp. e49–e57, Jul. 2006.

[12] A. Gretton, K. M. Borgwardt, M. Rasch, B. Schölkopf, and A. J. Smola, "A kernel method for the two-sample-problem," in *Proc. Adv. Neural Inf. Process. Syst.*, 2007, pp. 513–520.

[13] M. Long, Z. Cao, J. Wang, and M. I. Jordan, "Conditional adversarial domain adaptation," in *Proc. Adv. Neural Inf. Process. Syst.*, 2018, pp. 1640–1650.

[14] Z. Liu, Z. Miao, X. Zhan, J. Wang, B. Gong, and S. X. Yu, "Large-scale long-tailed recognition in an open world," in *Proc. IEEE/CVF Conf. Comput. Vis. Pattern Recognit.*, Jun. 2019, pp. 2537–2546.

[15] S. Sinha, H. Ohashi, and K. Nakamura, "Class-wise difficulty-balanced loss for solving class-imbalance," in *Proc. Asian Conf. Comput. Vis.*, 2020, pp. 1–17.

[16] Y. Hong, L. Niu, J. Zhang, W. Zhao, C. Fu, and L. Zhang, "F2GAN: Fusing-and-filling GAN for few-shot image generation," in *Proc. 28th ACM Int. Conf. Multimedia*, Oct. 2020, pp. 2535–2543.

[17] M. Ghifary, W. B. Kleijn, M. Zhang, D. Balduzzi, and W. Li, "Deep reconstruction-classification networks for unsupervised domain adaptation," in *Proc. ECCV*, 2016, pp. 597–613.

[18] J. Zhang, Z. Ding, W. Li, and P. Ogunbona, "Importance weighted adversarial nets for partial domain adaptation," in *Proc. IEEE Conf. Comput. Vis. Pattern Recognit.*, Jun. 2018, pp. 8156–8164.

[19] C.-Y. Lee, T. Batra, M. H. Baig, and D. Ulbricht, "Sliced wasserstein discrepancy for unsupervised domain adaptation," in *Proc. IEEE/CVF Conf. Comput. Vis. Pattern Recognit.*, Jun. 2019, pp. 10285–10295.

[20] Y. Zhang, H. Tang, K. Jia, and M. Tan, "Domain-symmetric networks for adversarial domain adaptation," in *Proc. IEEE/CVF Conf. Comput. Vis. Pattern Recognit. (CVPR)*, Jun. 2019, pp. 5031–5040.

[21] Y. Kim and S. Hong, "Adaptive graph adversarial networks for partial domain adaptation," *IEEE Trans. Circuits Syst. Video Technol.*, early access, Feb. 1, 2021, doi: 10.1109/TCSVT.2021.3056208.

[22] C. Schumann, X. Wang, A. Beutel, J. Chen, H. Qian, and E. H. Chi, "Transfer of machine learning fairness across domains," 2019, *arXiv:1906.09688*. [Online]. Available: http://arxiv.org/abs/1906.09688

[23] T. Jing and Z. Ding, "Adversarial dual distinct classifiers for unsupervised domain adaptation," in *Proc. IEEE Winter Conf. Appl. Comput. Vis. (WACV)*, Jan. 2021, pp. 605–614.

[24] S. J. Pan and Q. Yang, "A survey on transfer learning," *IEEE Trans. Knowl. Data Eng.*, vol. 22, no. 10, pp. 1345–1359, Oct. 2010.

[25] E. Tzeng, J. Hoffman, N. Zhang, K. Saenko, and T. Darrell, "Deep domain confusion: Maximizing for domain invariance," 2014, *arXiv:1412.3474*. [Online]. Available: http://arxiv.org/abs/1412.3474

[26] H. Xia and Z. Ding, "Structure preserving generative cross-domain learning," in *Proc. IEEE/CVF Conf. Comput. Vis. Pattern Recognit. (CVPR)*, Jun. 2020, pp. 4364–4373.

[27] H. Yan, Y. Ding, P. Li, Q. Wang, Y. Xu, and W. Zuo, "Mind the class weight bias: Weighted maximum mean discrepancy for unsupervised domain adaptation," in *Proc. IEEE Conf. Comput. Vis. Pattern Recognit. (CVPR)*, Jul. 2017, pp. 2272–2281.

[28] H. Xia and Z. Ding, "Hgnet: Hybrid generative network for zero-shot domain adaptation," in *Proc. 16th Eur. Conf. Comput. Vis. (ECCV)*, Glasgow, U.K. Springer, Aug. 2020, pp. 55–70.

[29] G. Kang, L. Jiang, Y. Yang, and A. G. Hauptmann, "Contrastive adaptation network for unsupervised domain adaptation," in *Proc. IEEE/CVF Conf. Comput. Vis. Pattern Recognit. (CVPR)*, Jun. 2019, pp. 4893–4902.

[30] J. Wang, Y. Chen, W. Feng, H. Yu, M. Huang, and Q. Yang, "Transfer learning with dynamic distribution adaptation," *ACM Trans. Intell. Syst. Technol.*, vol. 11, no. 1, pp. 1–25, Feb. 2020.

[31] M. Long, Y. Cao, J. Wang, and M. I. Jordan, "Learning transferable features with deep adaptation networks," 2015, *arXiv:1502.02791*. [Online]. Available: http://arxiv.org/abs/1502.02791

[32] J. Wang, W. Feng, Y. Chen, H. Yu, M. Huang, and P. S. Yu, "Visual domain adaptation with manifold embedded distribution alignment," in *Proc. 26th ACM Int. Conf. Multimedia*, 2018, pp. 402–410.

[33] J. Wang, Y. Chen, S. Hao, W. Feng, and Z. Shen, "Balanced distribution adaptation for transfer learning," in *Proc. IEEE Int. Conf. Data Mining (ICDM)*, Nov. 2017, pp. 1129–1134.

[34] E. Tzeng, J. Hoffman, K. Saenko, and T. Darrell, "Adversarial discriminative domain adaptation," in *Proc. IEEE Conf. Comput. Vis. Pattern Recognit.*, Jul. 2017, pp. 7167–7176.

[35] J. Shen, Y. Qu, W. Zhang, and Y. Yu, "Wasserstein distance guided representation learning for domain adaptation," in *Proc. 32nd AAAI Conf. Artif. Intell.*, Apr. 2018, pp. 4058–4065.

[36] T. Jing, H. Xia, and Z. Ding, "Adaptively-accumulated knowledge transfer for partial domain adaptation," in *Proc. 28th ACM Int. Conf. Multimedia*, Oct. 2020, pp. 1606–1614.

[37] Z. Cao, L. Ma, M. Long, and J. Wang, "Partial adversarial domain adaptation," in *Proc. Eur. Conf. Comput. Vis. (ECCV)*, 2018, pp. 135–150.

[38] T. Jing, H. Liu, and Z. Ding, "Towards novel target discovery through open-set domain adaptation," 2021, *arXiv:2105.02432*. [Online]. Available: http://arxiv.org/abs/2105.02432

[39] P. Panareda Busto and J. Gall, "Open set domain adaptation," in *Proc. IEEE Int. Conf. Comput. Vis.*, Oct. 2017, pp. 754–763.

[40] K. You, M. Long, Z. Cao, J. Wang, and M. I. Jordan, "Universal domain adaptation," in *Proc. IEEE/CVF Conf. Comput. Vis. Pattern Recognit.*, Jun. 2019, pp. 2720–2729.

[41] K. Saito, D. Kim, S. Sclaroff, and K. Saenko, "Universal domain adaptation through self supervision," in *Proc. Adv. Neural Inf. Process. Syst.*, vol. 33, 2020, pp. 1–14.

[42] T. M. H. Hsu, W. Y. Chen, C.-A. Hou, Y.-H.-H. Tsai, Y.-R. Yeh, and Y.-C.-F. Wang, "Unsupervised domain adaptation with imbalanced cross-domain data," in *Proc. IEEE Int. Conf. Comput. Vis. (ICCV)*, Dec. 2015, pp. 4121–4129.

[43] A. Beutel, J. Chen, Z. Zhao, and E. H. Chi, "Data decisions and theoretical implications when adversarially learning fair representations," 2017, *arXiv:1707.00075*. [Online]. Available: http://arxiv.org/abs/1707.00075

[44] A. Agarwal, A. Beygelzimer, M. Dudík, J. Langford, and H. Wallach, "A reductions approach to fair classification," 2018, *arXiv:1803.02453*. [Online]. Available: http://arxiv.org/abs/1803.02453

[45] A. Coston *et al.*, "Fair transfer learning with missing protected attributes," in *Proc. AAAI/ACM Conf. AI, Ethics, Soc.*, Jan. 2019, pp. 91–98.

[46] C. Lan and J. Huan, "Discriminatory transfer," 2017, *arXiv:1707.00780*. [Online]. Available: http://arxiv.org/abs/1707.00780

[47] K. He, X. Zhang, S. Ren, and J. Sun, "Deep residual learning for image recognition," in *Proc. IEEE Conf. Comput. Vis. Pattern Recognit.*, Jun. 2016, pp. 770–778.

[48] J. Deng, W. Dong, R. Socher, L.-J. Li, K. Li, and L. Fei-Fei, "Imagenet: A large-scale hierarchical image database," in *Proc. IEEE Conf. Comput. Vis. Pattern Recognit.*, Jun. 2009, pp. 248–255.

[49] P. Rodríguez, I. Laradji, A. Drouin, and A. Lacoste, "Embedding propagation: Smoother manifold for few-shot classification," in *Proc. Eur. Conf. Comput. Vis.* Cham, Switzerland: Springer, 2020, pp. 121–138.

[50] R. Soentpiet *et al.*, *Advances in Kernel Methods: Support Vector Learning*. Cambridge, MA, USA: MIT Press, 1999.

[51] W. S. Lee, P. L. Bartlett, and R. C. Williamson, "Lower bounds on the VC dimension of smoothly parameterized function classes," *Neural Comput.*, vol. 7, no. 5, pp. 1040–1053, Sep. 1995.

[52] Y. Liu *et al.*, "Learning to propagate labels: Transductive propagation network for few-shot learning," 2018, *arXiv:1805.10002*. [Online]. Available: http://arxiv.org/abs/1805.10002

[53] D. Zhou, O. Bousquet, T. N. Lal, J. Weston, and B. Schölkopf, "Learning with local and global consistency," in *Proc. Adv. Neural Inf. Process. Syst.*, vol. 16, 2004, pp. 321–328.

[54] P. Upchurch *et al.*, "Deep feature interpolation for image content changes," in *Proc. IEEE Conf. Comput. Vis. pattern Recognit.*, Oct. 2017, pp. 7064–7073.

[55] M. Xu *et al.*, "Adversarial domain adaptation with domain mixup," 2019, *arXiv:1912.01805*. [Online]. Available: http://arxiv.org/abs/1912.01805

[56] S. Motiian, Q. Jones, S. Iranmanesh, and G. Doretto, "Few-shot adversarial domain adaptation," in *Proc. Adv. Neural Inf. Process. Syst.*, 2017, pp. 6670–6680.

[57] J. Snell, K. Swersky, and R. Zemel, "Prototypical networks for few-shot learning," in *Proc. Adv. Neural Inf. Process. Syst.*, 2017, pp. 4077–4087.

[58] K. Saenko, B. Kulis, M. Fritz, and T. Darrell, "Adapting visual category models to new domains," in *Proc. Eur. Conf. Comput. Vis.* Berlin, Germany: Springer, 2010, pp. 213–226.

[59] H. Venkateswara, J. Eusebio, S. Chakraborty, and S. Panchanathan, "Deep hashing network for unsupervised domain adaptation," in *Proc. IEEE Conf. Comput. Vis. Pattern Recognit.*, Jul. 2017, pp. 5018–5027.

[60] Y. LeCun, L. Bottou, Y. Bengio, and P. Haffner, "Gradient-based learning applied to document recognition," *Proc. IEEE*, vol. 86, no. 11, pp. 2278–2324, Nov. 1998.

[61] C. Szegedy *et al.*, "Going deeper with convolutions," in *Proc. IEEE Conf. Comput. Vis. Pattern Recognit. (CVPR)*, Jun. 2015, pp. 1–9.

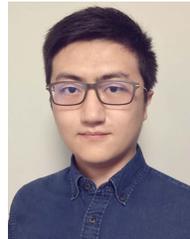

**Taotao Jing** received the B.S. degree in electronic science and technology from Xi'an Jiaotong University, Xi'an, China, in 2016, and the M.S. degree in computer system engineering from Northeastern University, Boston, MA, USA, in 2018. He is currently pursuing the Ph.D. degree with the Department of Computer Science, Tulane University. His research interests lie on computer vision, transfer learning, and deep learning.

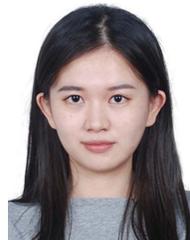

**Bingrong Xu** received the B.S. degree from the School of Automation, Wuhan University of Technology, Wuhan, China, in 2015. She is currently pursuing the Ph.D. degree with the School of Artificial Intelligence and Automation, Huazhong University of Science and Technology, Wuhan. Her current research interests include zero-shot learning, transfer learning, sparse representation, and low-rank representation.

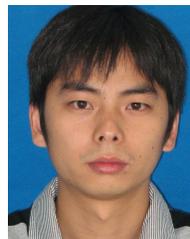

**Zhengming Ding** (Member, IEEE) received the B.Eng. degree in information security and the M.Eng. degree in computer software and theory from the University of Electronic Science and Technology of China (UESTC), China, in 2010 and 2013, respectively, and the Ph.D. degree from the Department of Electrical and Computer Engineering, Northeastern University, USA, in 2018. He has been a Faculty Member affiliated with the Department of Computer Science, Tulane University, since 2021. Prior that, he was a Faculty Member affiliated with the Department of Computer, Information and Technology, Indiana University–Purdue University Indianapolis. His research interests include transfer learning, multi-view learning, and deep learning. He received the National Institute of Justice Fellowship, from 2016 to 2018. He was a recipient of the Best Paper Award (SPIE 2016) and Best Paper Candidate (ACM MM 2017). He is currently an Associate Editor of the *Journal of Electronic Imaging* (JEI) and *IET Image Processing*. He is a member of the ACM and AAAI.